\documentclass[lettersize,journal]{IEEEtran}
\usepackage{amsmath,amsfonts}
\usepackage[ruled,vlined]{algorithm2e}
\usepackage{array}
\usepackage[caption=false,font=normalsize]{subfig}
\usepackage{textcomp}
\usepackage{stfloats}
\usepackage{url}
\usepackage{verbatim}
\usepackage{graphicx}
\usepackage{cite}
\usepackage{tcolorbox}
\usepackage{multirow}

\usepackage{framed}
\definecolor{darkgreen}{HTML}{006400}

\begin{document}

\newcommand{\xl}[1]{{\color{orange}{[xl:#1]}}}
\newcommand{\fei}[1]{{\color{purple}{[fei:#1]}}}

\title{Algorithm Evolution Using Large Language Model }

\author{Fei Liu, Xialiang Tong, Mingxuan Yuan and Qingfu Zhang,~\IEEEmembership{Fellow,~IEEE}
\IEEEcompsocitemizethanks{\IEEEcompsocthanksitem 

Fei Liu and Qingfu Zhang are with the Department of Computer Science, City University of Hong Kong, Hong Kong (e-mail: fliu36-c@my.cityu.edu.hk; qingfu.zhang@cityu.edu.hk).

Xialiang Tong and Mingxuan Yuan are with Huawei Noah’s Ark Lab (e-mail: tongxialiang@huawei.com;
yuan.mingxuan@huawei.com). 
}
}


\markboth{IEEE Journal}%
{Shell \MakeLowercase{\textit{et al.}}: A Sample Article Using IEEEtran.cls for IEEE Journals}


\maketitle

\begin{abstract}

Optimization can be found in many real-life applications. Designing an 
effective algorithm for a specific optimization problem typically requires a tedious amount of effort from human experts with domain knowledge and algorithm design skills. In this paper, we propose a novel approach called Algorithm Evolution using Large Language Model (AEL). It utilizes a large language model (LLM) to automatically generate optimization algorithms via an evolutionary framework. AEL does algorithm-level evolution without model training. Human effort and requirements for domain knowledge can be significantly reduced. We take constructive methods for the traveling salesman problem as a test example, we show that the constructive algorithm obtained by AEL outperforms simple hand-crafted and LLM-generated heuristics. Compared with other domain deep learning model-based algorithms, these methods exhibit excellent scalability across different problem sizes. AEL is also very different from previous attempts that utilize LLMs as search operators in algorithms. 
\end{abstract}


\section{Introduction}

\IEEEPARstart{O}{ptimization} is everywhere. It plays a crucial role in production, planning, decision making, and resource management. Numerous research works have been carried out to develop efficient and powerful optimization algorithms. While these algorithms have proven to be valuable tools in many practical scenarios, designing them usually requires extensive manual crafting with domain knowledge~\cite{nocedal1999numerical,glover2006handbook,deb2012optimization}.

\begin{figure}[t]
    \centering
    \subfloat{\includegraphics[width=0.9\linewidth]{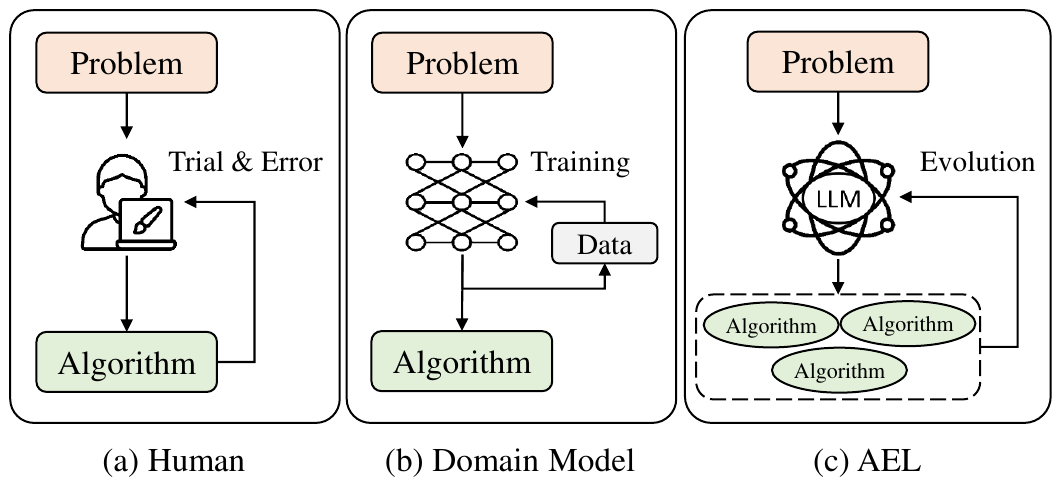}} \\
    \subfloat{\includegraphics[width=0.95\linewidth]{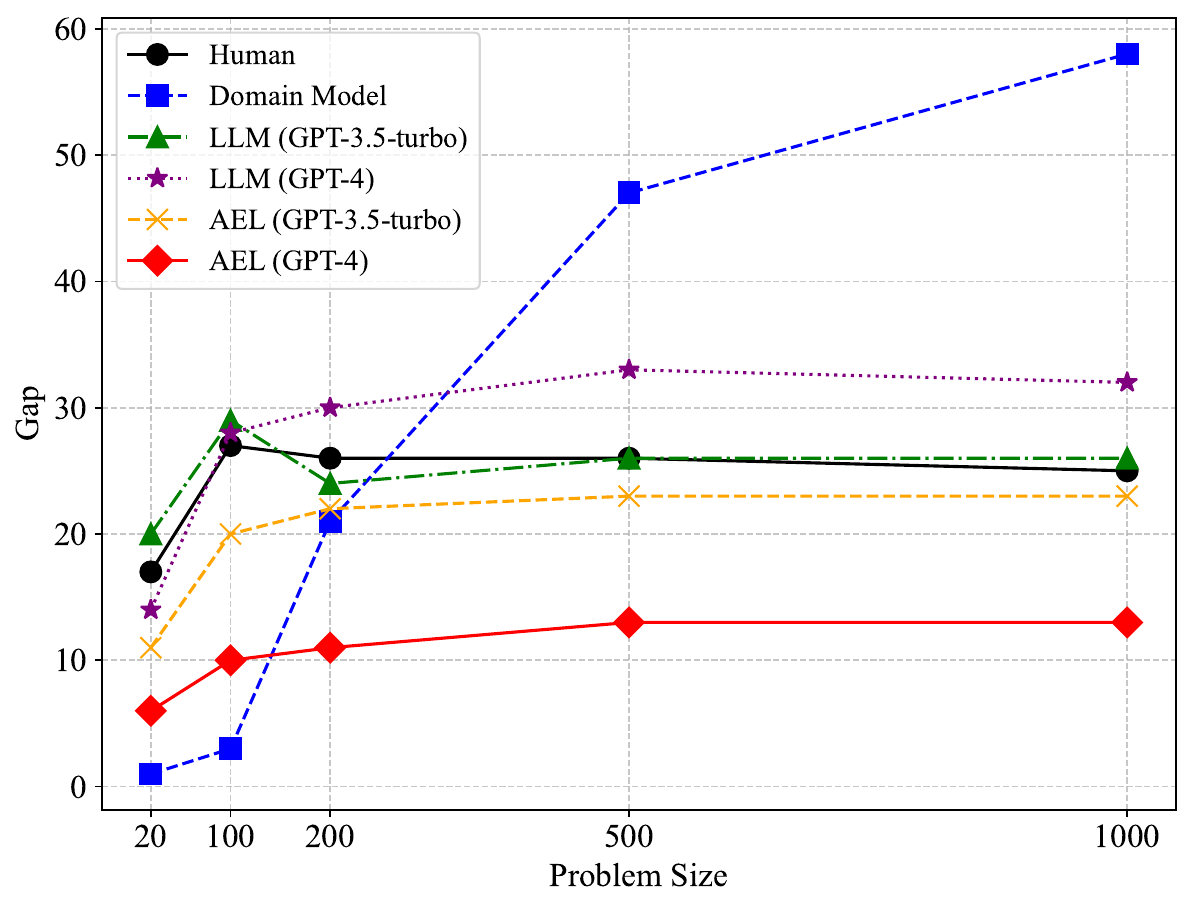}}
    \caption{A comparison of three different algorithm design approaches \textbf{(a) Human}, \textbf{(b) Domain Model}, and \textbf{(c) AEL}, and the their results on TSP. The x-axis represents the problem size. The y-axis represents the gap (\%) to the baseline. All results are averaged on 64 randomly generated instances. \textbf{(a) Human (Greedy):} an algorithm designed by humans with trial and error (a greedy algorithm). \textbf{(b) Domain Model:} an algorithm learned by a specific deep neural network trained on TSP50. \textbf{(c) AEL:} algorithms created by our proposed AEL evolved on TSP50. We also compare the algorithms directly generated by instructing LLM (\textbf{LLM}). The used LLMs are denoted in brackets. Refer to the experimental section for more details.}
    \label{fig:comparison}
\end{figure}

To overcome these limitations, researchers have made much effort to automate the algorithm design process. Learning to optimize~\cite{chen2022learning} involves automatically designing optimization methods based on their performance on a training set of problems. In deep learning, AutoML~\cite{he2021automl} offers promising solutions for constructing a deep learning system without human intervention. In the context of heuristic design, this is commonly referred to as hyperheuristics~\cite{burke2013hyper,burke2019classification} or automatic design of heuristics~\cite{stutzle2019automated}. Moreover, reinforcement learning~\cite{ma2021learning,tian2022deep,zhang2023reinforcement,wu2021learning}, supervised learning~\cite{shen2023adaptive,sun2022boosting}, transfer learning~\cite{tan2021evolutionary,zhou2020toward,li2023data} and meta-learning~\cite{liu2021prediction,zhang2022meta} have also been widely employed to enhance optimization efficiency, promote algorithm performance, and generate new algorithms. Furthermore, recent works have explored the use of end-to-end neural solvers for both continuous optimization~\cite{cao2019learning,lange2023discovering,penghui2022decn,li2022optformer,jiang2023knowledge} and combinatorial optimization~\cite{kool2018attention,lin2022pareto_combinatorial,zhou2023towards,luo2023neural,wang2023multiobjective}. While these progresses have significantly advanced the field of automatic algorithm design, they usually require an underlying algorithm framework or a time-consuming domain model crafting and training.

In the past three years, large language models (LLMs) have demonstrated remarkable capabilities in various research domains~\cite{zhao2023survey,kasneci2023chatgpt}, such as natural language processing~\cite{min2021recent}, programming~\cite{tian2023chatgpt}, medicine~\cite{lee2023benefits,nori2023capabilities,cheng2023exploring}, chemistry~\cite{jablonka2023gpt}, chip design~\cite{blocklove2023chip,he2023chateda}, and optimization~\cite{yu2023gpt,zheng2023can,zhang2023automl,zhao2023large}. Recently, several works have adopted LLMs as pre-trained black-box optimizers for optimization~\cite{yang2023large,meyerson2023language,lange2023discovering,chen2023evoprompting,wang2023can}. However, these works use LLMs to generate new solutions at the operator level. Its performance declines considerably when applied to large-scale problems, mainly due to the longer solution representation and large search space~\cite{yang2023large,liu2023large,wang2023can}.


In this paper, we propose a novel approach for practical automatic algorithm design called \textbf{Algorithm Evolution using Large Language Model (AEL)}. AEL distinguishes itself from algorithm design by humans and through training domain models. It creates and modifies algorithms by interacting with large language models within an evolutionary framework. We demonstrate the effectiveness of AEL on the constructive method for the traveling salesman problem (TSP), highlighting its ability to generate novel and scalable algorithms. The contributions of this paper are summarized as follows:
\begin{itemize}
    \item  We introduce AEL, which treats each algorithm as an individual and utilizes an evolutionary framework to evolve new algorithms. Our approach integrates large language models at the algorithm design level, allowing for the creation and modification of existing search strategies with minimal expert knowledge and no domain model training.
    \item We demonstrate AEL on designing the constructive heuristic for TSP, a well-known combinatorial optimization problem. The individual representation and prompt engineering strategies for TSP are designed.
    \item We compare three algorithm design approaches (a) a greedy algorithm designed by humans (\textbf{Human}), (b) an algorithm learned by a specific model through time-consuming training (\textbf{Domain Model}), and (c) algorithms designed automatically by our proposed AEL (\textbf{AEL)}. The comparison of these three approaches and a summary of results on TSP with different problem sizes are illustrated in Fig.~\ref{fig:comparison}. The results demonstrate that AEL outperforms the manually designed greedy heuristic for all problem sizes. Furthermore, AEL exhibits better generalization performance compared to training a domain model. We also highlight the superiority of AEL over directly instructing LLM (\textbf{LLM}) to generate algorithms. AEL benefits from the use of more advanced language models, such as GPT-4.
\end{itemize}

The rest of this paper is structured as follows: In Section~\ref{sec:relate}, we provide an overview of related work in automatic algorithm design and the use of large language models in optimization. Section~\ref{sec:ael} presents the framework and methodology of our proposed AEL. In Section~\ref{sec:tsp}, we present the demonstration and experimental results on TSP and a discussion of findings followed by some suggested future directions. Section~\ref{sec:conclusion} concludes the paper.

\section{Related Works}\label{sec:relate}

\subsection{Automatic Algorithm Design (AAD)}
Automatic algorithm design (AAD) is an active and rapidly growing research area in heuristic design~\cite{gendreau2010handbook}. It is commonly referred to as hyper-heuristics~\cite{burke2013hyper,burke2019classification} or automatic design of heuristics~\cite{stutzle2019automated}. A number of effective toolboxes and frameworks have been proposed, such as GGA~\cite{ansotegui2009gender}, ParamILS~\cite{hutter2009paramils}, MO-ParamILS~\cite{blot2016mo}, irace~\cite{lopez2016irace}, SMAC~\cite{hutter2011sequential}, and Optuna~\cite{akiba2019optuna}, which significantly facilitate researchers in the field. From the algorithm perspective, AAD can be briefly categorized into three main approaches: automatic algorithm configuration, automatic algorithm selection, and automatic algorithm composition~\cite{meng2021automated}. Many recent research works in AAD focus on automatically generating improved algorithms using general heuristic components~\cite{stutzle2019automated}. For example, AutoEMOA for multi-objective evolutionary optimization~\cite{bezerra2015automatic} and AutoGCOP for general-purpose optimization~\cite{meng2021automated}. Despite these advancements, these approaches still require a backbone domain algorithm framework or a set of manually crafted algorithm components, and the designing process can be time-consuming.

\subsection{Machine Learning for Optimization}
Over the past few decades, much effort has been made on the integration of machine learning techniques for optimization~\cite{bengio2021machine,chen2022learning,he2021automl,li2023survey}. Reinforcement learning techniques have been used to learn optimal algorithm configurations~\cite{ma2021learning}, policy for operator selection~\cite{tian2022deep,zhang2023reinforcement}, solution selection~\cite{wu2021learning}, and construction of the partial solution~\cite{liu2022hybridization}. \cite{dong2016supervised,shen2023adaptive,sun2022boosting,sun2022learning} adopt supervised learning for solution prediction to boost heuristics and mathematical programming. Transfer learning~\cite{tan2021evolutionary,zhou2020toward} and meta-learning~\cite{liu2021prediction,zhang2022meta} have also been employed to transfer and extract knowledge from different tasks or solutions to improve optimization efficiency. Other popular directions include using surrogate models to approximate the objective functions or relations, which have been successfully applied in expensive optimization in various domains~\cite{shahriari2015taking,zhang2009expensive,jin2018data,song2021kriging,hao2022expensive}, and learning through genetic programming~\cite{mei2022explainable,jia2022learning}, which is explainable and well-suited for structure representation.

Furthermore, recent works have explored the use of end-to-end neural solvers for both continuous optimization~\cite{cao2019learning,lange2023discovering,penghui2022decn,li2022optformer,jiang2023knowledge} and combinatorial optimization~\cite{kool2018attention,lin2022pareto_combinatorial,zhou2023towards,luo2023neural,wang2023multiobjective}. Some of them have been applied for EC. For example, \cite{cao2019learning,lange2023discovering,penghui2022decn,li2022optformer} train deep neural networks to approximate mutation, crossover, and evolutionary strategies for black-box optimization. \cite{jiang2023knowledge} enhances evolutionary algorithms through learning from historically successful experiences. However, in spite of the promising results, they often require significant effort in designing and training the domain models.

\subsection{Large Language Model (LLM)}
In the last three years, large language models (LLMs) have gained increasing power due to their exponentially growing model sizes and the availability of large training datasets. LLMs have shown remarkable performance in various research domains including natural language processing~\cite{min2021recent}, programming~\cite{tian2023chatgpt}, medicine~\cite{lee2023benefits,nori2023capabilities,cheng2023exploring}, chemistry~\cite{jablonka2023gpt}, chip design~\cite{blocklove2023chip,he2023chateda}, and optimization~\cite{yu2023gpt,zheng2023can,zhang2023automl}. These LLMs excel at performing diverse tasks in a zero-shot manner~\cite{zhao2023survey,kasneci2023chatgpt}.  Despite these advancements, the practical utilization of LLMs for designing optimization algorithms is still in its early stages.

\subsection{LLMs for Algorithm Design}
Recently, several works have demonstrated the potential of optimization solely through prompting LLM~\cite{yang2023large,guo2023towards}. \cite{liu2023large} proposes a decomposition-based framework that integrates LLM as a black-box operator for multiobjective evolutionary optimization. In the context of single-objective evolutionary algorithms, \cite{yang2023large,brownlee2023enhancing,liu22023large} adopt LLM for the selection, crossover, and mutation processes. LLM has also been integrated into neural architecture design~\cite{nasir2023llmatic,jawahar2023llm,chen2023evoprompting}, Monte Carlo Tree Search~\cite{zhao2023large} and graph-based combinatorial optimization~\cite{wang2023can}. The applications of LLM for genetic programming and open-ended tasks are discussed in \cite{lehman2023evolution, meyerson2023language}.

However, most of these works directly use LLM as optimizers, which suffers from poor generalization performance on large-scale problems. The longer solution representation and large search space, especially on combinatorial optimization problems~\cite{yang2023large,liu22023large,wang2023can}, pose significant challenges to zero-shot generalization and in-context learning for LLM. \cite{liu2023large} proposes to use an explainable lightweight operator to approximate the results of LLMs for better generalization in continuous optimization, which is not suitable for addressing combinatorial optimization problems.

\section{Algorithm Evolution using Large Language Model}\label{sec:ael}
\begin{figure*}[htbp]
    \centering
    \includegraphics[width=0.9\textwidth]{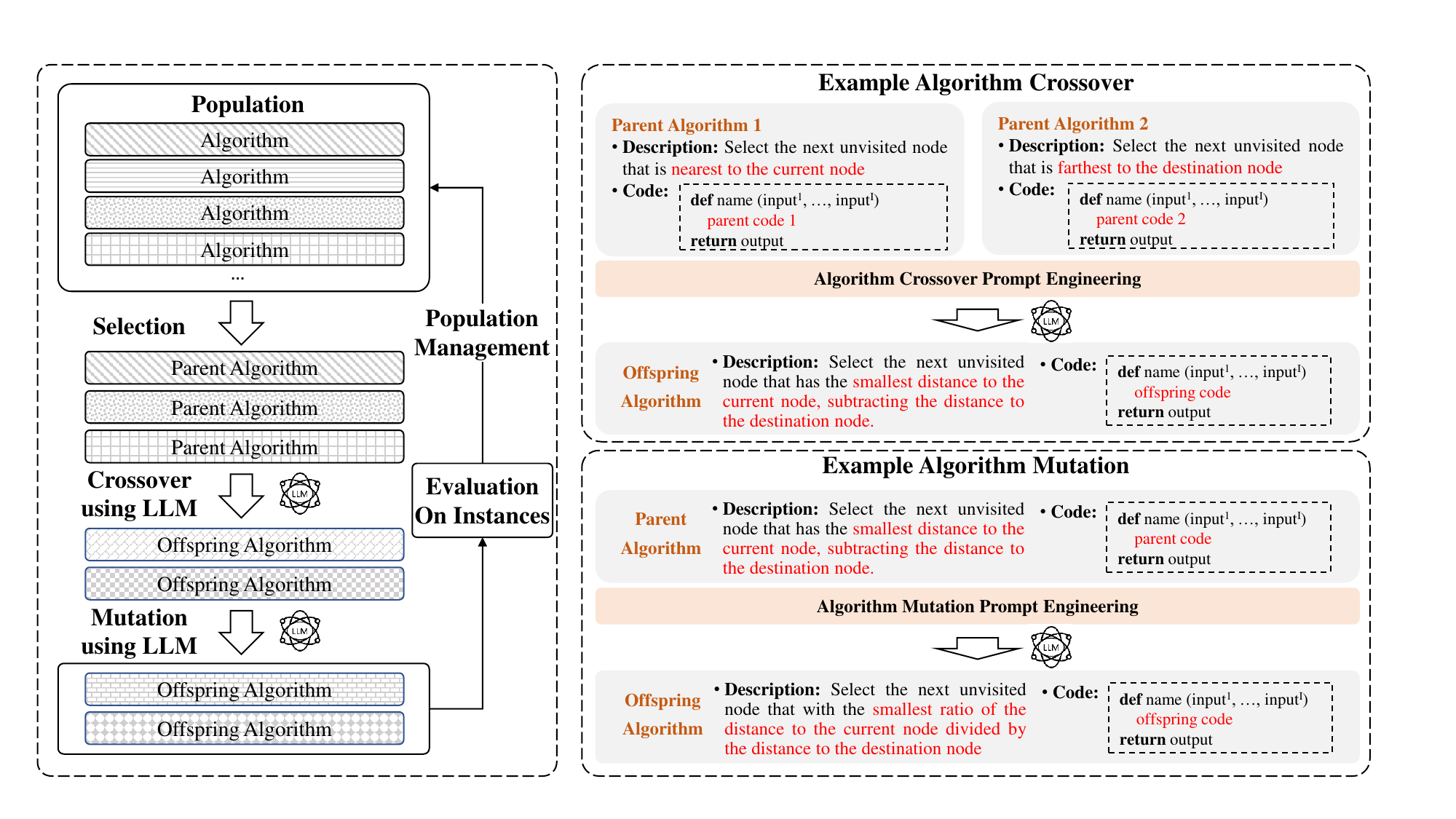}
    \caption{An illustration of the AEL framework. The left-hand side flowchart adopts a standard evolutionary framework, comprising prompt engineering of LLM for initialization, crossover, and mutation to create/evolve new algorithms. On the right-hand side, there are two examples demonstrating algorithm crossover and algorithm mutation, specifically in their application to selecting the next node in a route. }
    \label{fig:framework}
\end{figure*}

\subsection{AEL Framework}

The proposed Algorithm Evolutionary using Large Language Model (AEL) framework embraces the common framework utilized in evolutionary computing (EC). It evolves a population of individuals and comprises fundamental components, including initialization, selection, crossover, mutation, and population management. In spite of the general evolution framework, AEL differs significantly from existing approaches. 

\begin{itemize}
    \item Firstly, unlike major algorithms in EC, where individuals represent feasible solutions, each individual within our AEL framework represents an algorithm designed explicitly for a given problem. AEL evolves algorithms capable of generating novel and competitive search strategies for a target problem, rather than seeking improved solutions for specific instances.
    \item Furthermore, AEL distinguishes itself from other automatic algorithm design methods. By integrating LLMs into an evolutionary framework, the creation and refinement of algorithms occur automatically, eliminating the need for training new models or utilizing baseline algorithms. In contrast, existing automatic algorithm design methods often require expensive searches for improved algorithms or rely on specific model training.
\end{itemize}

Fig.~\ref{fig:framework} illustrates the AEL flowchart and examples of algorithm crossover and algorithm mutation. AEL begins with an initialization step, where a population of $N$ individuals (algorithms), denoted as $P = \{a_1, \dots, a_N\}$, is created and evaluated. The creation of algorithms in the initializations can be either using existing algorithms or letting LLM generate algorithms. Each individual $a_j$ is evaluated using a fitness function, resulting in the fitness value $f(a_j)$. The fitness is evaluated on a branch of evaluation instances instead of on one instance.

The framework then proceeds with a series of iterations, for a total of $N_g$ generations with $N$ iterations for each generation. In each iteration, a subset of $l$ individuals $p_j = \{a_1, \dots, a_l\}$ is selected from the population using a selection method (we select each individual with equal probability in the experiments). The selected individuals are then subjected to a crossover operation with a probability $\theta_1$, resulting in a new set of $s$ individuals $o_j = \{a_1, \dots, a_s\}$ created by LLM. The crossover operation ensures the exploration of different genetic information from the input subset $p_j$. 

For each individual $a_k$ in the set $o_j$, a mutation operation is applied with a probability $\theta_2$. This operation modifies the input algorithm, potentially introducing a new algorithm or an algorithm with different parameters into the population. Subsequently, the fitness of each mutated individual $a_k$ is evaluated.

To maintain a constant population size during the evolution process, population management is performed. The new individuals $o_j$ are added to the population $P$. Afterward, the population $P$ is managed to reduce its size from $(s+1) \times N$ to $N$ by deleting the worst ones.

Ultimately, the AEL Framework aims to find the best algorithm within the given population, denoted as $a^*$. By iteratively applying selection, crossover, mutation, and population management, AEL stimulates LLM to evolve a better algorithm for the optimization problem at hand.

\begin{algorithm}[ht]
    \caption{AEL Framework}
    \label{alg:framework}
    \KwIn{The number of population: $N_g$; Population size $N$; Probability of crossover $\sigma_1$; Probability of mutation $\sigma_2$; The number of parents $l$ for crossover; The number of new individuals $s$ for crossover; A given LLM.}
    \KwOut{Best algorithm $a^*$.}
    \textbf{Initialization:} 

    \For{$j=1,\dots,N$}{
        \textbf{Algorithm Creation:} create new individual $a_j$ given the target problem using LLM;
        Evaluate $a_j$ and get fitness value $f(a_j)$;
    }
    
    Construct initial population $P=\{a_1,\dots,a_N\}$;
    
    \For{$i=1,\dots,N_g$}{
        \For {$j=1,\dots,N$}{
            \textbf{Selection:} select a subset of input individuals $p_j=\{a_1,\dots,a_l \}$;

            \textbf{Algorithm Crossover} with probability $\theta_1$: create individual $o_j=\{a_1,...,a_s\}$ using LLM given the target problem and input subset $p_j$;

            \For{$k=1,\dots,s$}{
                \textbf{Algorithm Mutation} with probability $\theta_2$: modify individual $a_k$ using LLM;
                
                Evaluate $a_k$ and get fitness value $f(a_k)$;
            }
        }
            
        \textbf{Population management:} $P=P\cup \{o_1,...,o_N\}$, manage population $P$ to reduce the size from $(s+1)\dot N$ to $N$.
        }
\end{algorithm}

\subsection{Individual Representation}
Unlike previous works, our proposed AEL approach represents each individual as an algorithm tailored to the specific problem. Each individual consists of three components: 1) a description of the algorithm in natural language, 2) a code block in a pre-defined format, and 3) a fitness value.

The algorithm description comprises a few sentences in natural language that can be easily processed by LLM. The code block should follow a predefined format so that it can be identified and seamlessly integrated into our AEL framework. We introduce the four basic components that should be included in the prompt engineering to format the code block:
\begin{itemize}
    \item Name of class or function: The name of the class or function must be standardized to ensure easy identification by the main program.
    \item Input: The number and names of input variables need to be provided, along with their types and their characters in the program. This information helps LLM understand the available information and design a viable algorithm based on it.
    \item Output: The number and names of output variables should also be defined, along with their types and their utilization.
    \item Other hints: we expect the response to be innovative, and want to avoid too many explanations and code comments, which might lead to failure identification and increase the response time and cost. Any other problem-specific hints should also be included in the prompt.
\end{itemize}

The evaluation of individuals in AEL involves running the algorithms on an evaluation instance set of the target problem. This evaluation process differs from traditional evolutionary computing, which typically evaluates the objective function for a single instance, but is closer to AAD methods~\cite{hutter2009paramils,lopez2016irace,hutter2011sequential}.

\subsection{Initilization}
The initial population can be either constructed by using existing manually crafted algorithms or created using LLM. In our experiments, we choose to let LLM create all the initial algorithms to eliminate the use of expert knowledge. We provide the following guidelines for prompt engineering for algorithm creation using LLMs. The prompt should include the following three parts:
\begin{itemize}
\item \textbf{A description of the task:} A description of the optimization task. The description should be a concise but comprehensive introduction to the target problem.
\item \textbf{An expected output:} A description of the expected responses that we want. We desire both a description of the new algorithm and a corresponding code block with pre-defined inputs and outputs. The responses should be provided in a specific format.
\item \textbf{Initialization-specific hints:} We prioritize the inclusion of diverse algorithms during the initialization process, thus emphasizing the development of a novel algorithm that distinguishes itself from those in the literature. We may also encourage more randomness and diversity by explicitly instructing the LLM to be creative.
\end{itemize}
We run the prompting procedure for $N$ times to generate $N$ initial algorithms.

\subsection{Selection}
We simply select input $l$ algorithms randomly from the population $P$. This selection strategy aligns with the conventional EC. The difference is that The number of selected individuals in our framework is scalable, as the LLM takes prompt in a flexible manner. This scalability is only limited by the maximum input token size depending on the adopted LLM.

\subsection{Crossover}
For the $j$-th iteration, the input of crossover consists of a set of $l$ selected individuals, denoted as $p_j = \{a_1, \dots, a_l\}$, and the output is a new set of $s$ individuals, denoted as $o_j = \{a_1, \dots, a_s\}$. The process of prompt engineering for crossover involves using the target problem and the selected individuals to generate a prompt for LLM, which in turn creates new algorithms. The prompt itself is a combination of natural language and code and is structured into four parts:

\begin{itemize}
    \item \textbf{A description of the task: } It is identical to that used during initialization.
    \item \textbf{Parent algorithms:} It comprises a set of $l$ input parent individuals. For each individual, both the algorithm description and code are provided.
    \item \textbf{An expected output:} It includes a description of the desired responses in a specific format.
    \item \textbf{Crossover-specific hints: } It emphasizes the need for the new algorithm to draw motivation from the parent algorithms while still being distinct from them.
\end{itemize}

\subsection{Mutation}
In mutation, the input algorithm is modified using LLM to create a new algorithm. The engineering approach for mutation involves using the target problem and one input individual to generate a prompt that allows for the modification of the input algorithm with LLM. The prompt consists of a combination of natural language and code, and it includes the following four components:
\begin{itemize}
    \item \textbf{A description of the task:} It is identical to that used during initialization.
    \item \textbf{A parent algorithm:} Only one input parent individual. Both the algorithm description and code are provided. 
    \item \textbf{An expected output:} A description of the expected responses we want. The responses should be given in a specific format.
    \item \textbf{Mutation-specific hints:} It emphasizes that the new algorithm should be a revision of the input algorithm.
\end{itemize}

\subsection{Population Management}
In population management, we remove the worst individuals in terms of fitness value to reduce the population size from $(s+1)N$ to $N$.

\section{Demonstration on TSP}\label{sec:tsp}
The traveling salesman problem (TSP) is one of the most important combinatorial optimization problems. The problem is to find the shortest route to visit all the given locations once and return to the starting location. It is recognized as NP-hard and is usually solved using heuristic algorithms. Among different heuristics~\cite{reinelt2003traveling}, constructive heuristics are flexible and easy to implement. In constructive heuristics, the solution is constructed step-by-step by choosing the next node given the current node and the destination node. This autoregressive manner is also used by many recent works on learning a domain neural model for combinatorial optimization~\cite{kool2018attention,kwon2020pomo}. 

We adopt the same constructive framework to iteratively choose the next node. The task for AEL in this case is to create a novel and competitive algorithm to choose the next node.

\subsection{AEL Implementation}

AEL is a general framework for algorithm design. For a target problem like TSP, we only need to implement the problem-specific parts. i.e., the individual representation and the prompting engineering for initialization, crossover, and mutation.

\subsubsection{Individual Representation}
\begin{figure}[htbp]
    \centering
    \includegraphics[width=\linewidth]{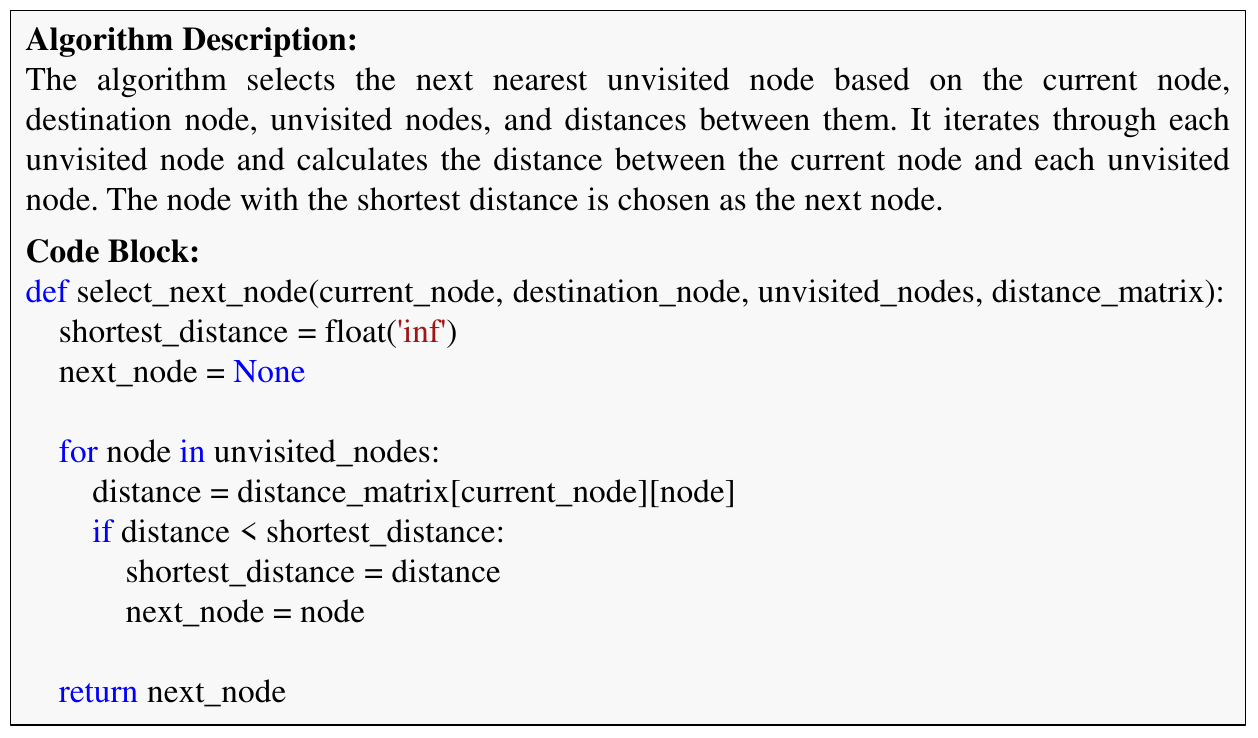}
    \caption{An example of individual representation of the greedy algorithm. \textbf{Algorithm Description} is a brief algorithm description in two sentences. \textbf{Code Block} includes a Python function named "select\_next\_node" with a pre-defined input and output. The fitness value is a real number, which is not depicted.}
    \label{fig:greedy}
\end{figure}

The objective of AEL is to evolve an algorithm that identifies the next node based on problem information and the current state. As mentioned previously, each individual in AEL comprises three components: 1) an algorithm description, 2) a code block, and 3) a fitness value. Fig~\ref{fig:greedy} presents an example of the individual representation of the greedy algorithm, which selects the next node that is the nearest one to the current node.

The algorithm description is presented in a natural language format. To avoid a noisy long response, we explicitly inform the LLM that the algorithm description should not exceed two sentences. It is important to note that, in more complex tasks, a longer sentence limitation can be adopted. In addition to it, we pose no additional limitations. 

The code is a Python function named "select\_next\_node", which takes inputs including the current node, destination node, unvisited nodes, and distance matrix, and outputs the next selected node. 

The fitness value is evaluated using a set of 64 randomly generated TSP instances of size 50. The fitness is calculated as the average gap to the commercial solver Gurobi. Smaller values indicate better fitness.

\subsubsection{Prompt Engineering for Initialization, Crossover, and Mutation}
\begin{figure}[htbp]
    \centering
    \subfloat{\includegraphics[width=0.8\linewidth]{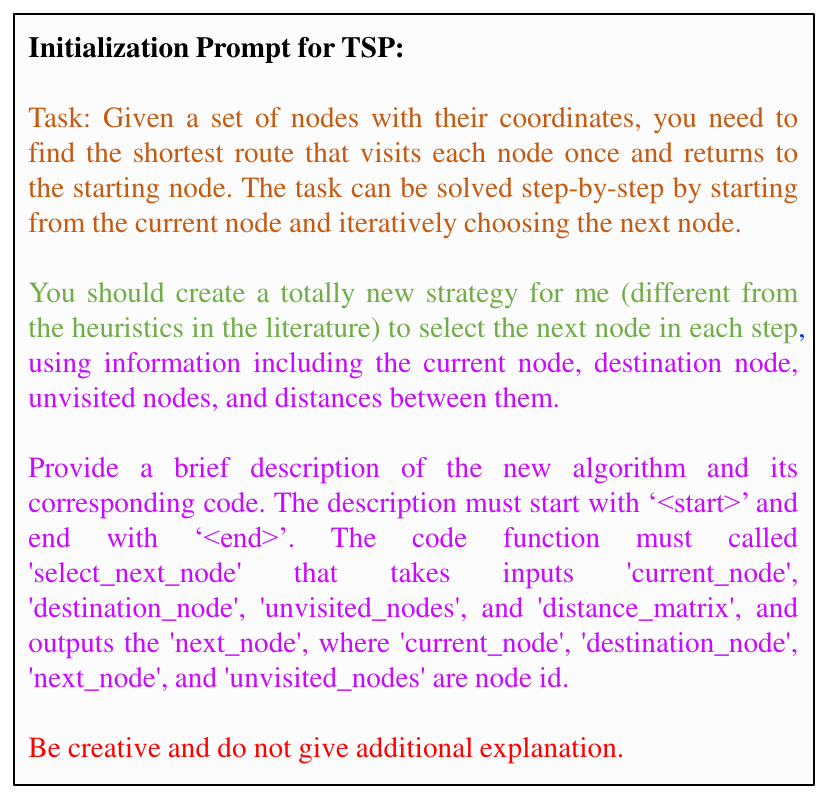}}
    \\
    \subfloat{\includegraphics[width=0.8\linewidth]{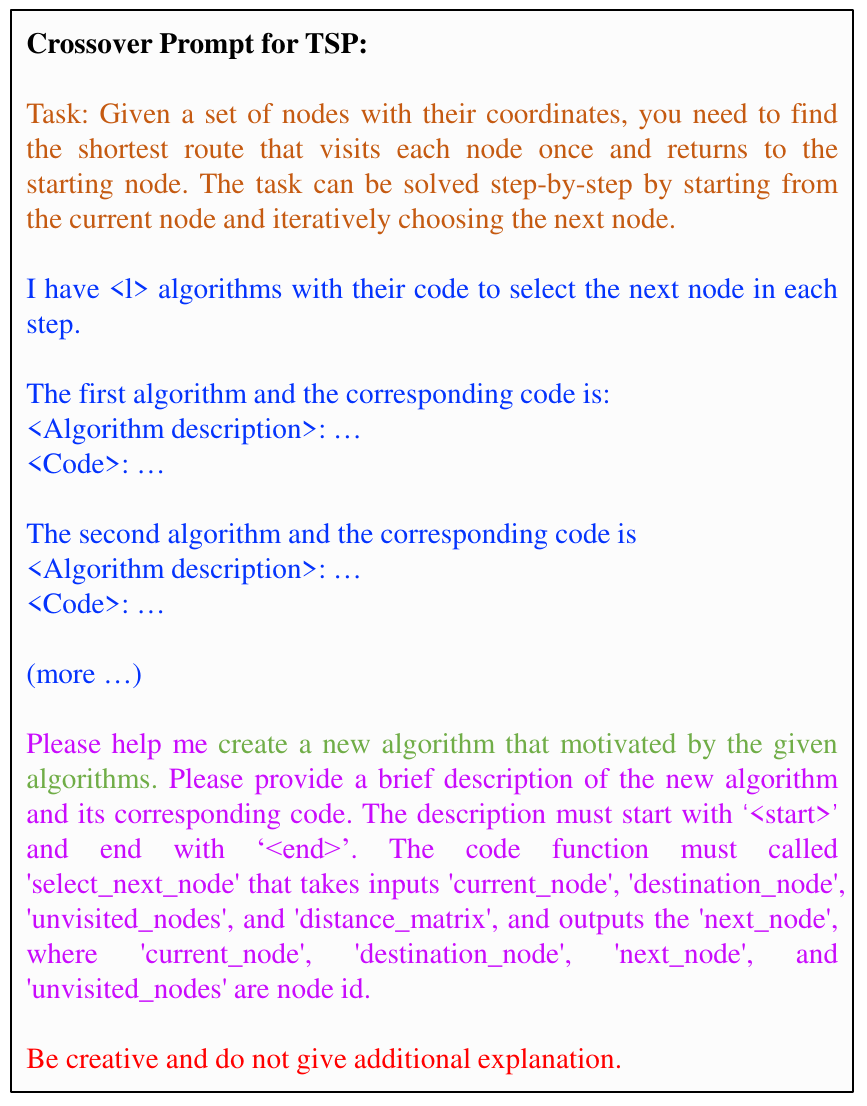}}
    \\
    \subfloat{\includegraphics[width=0.8\linewidth]{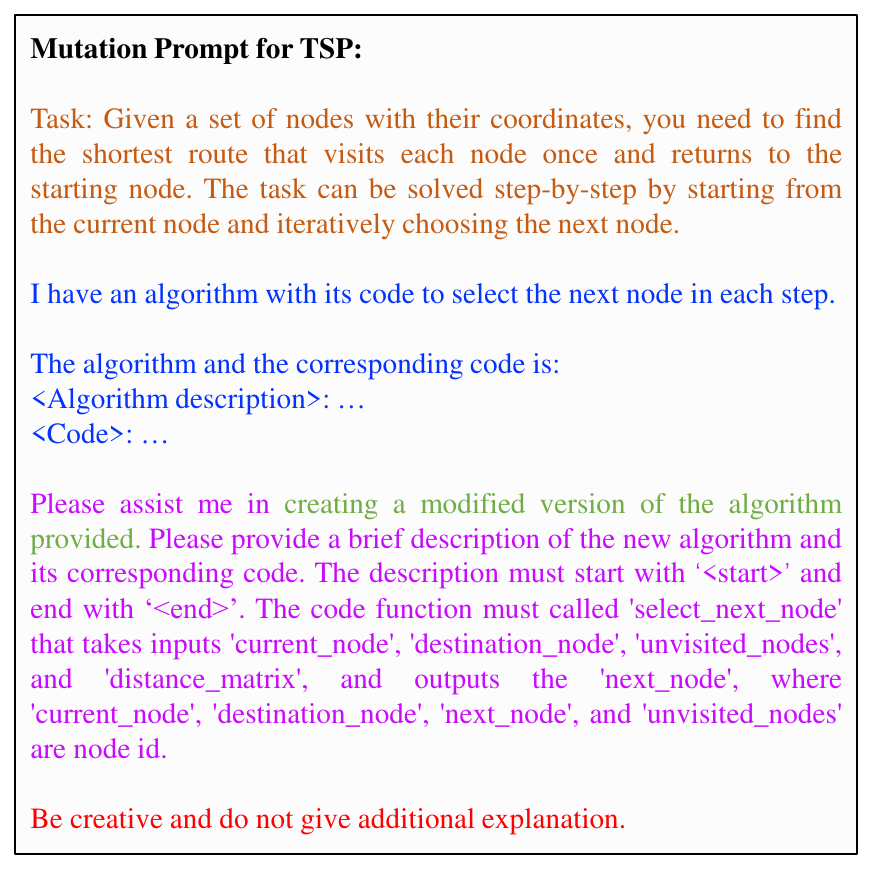}}
    \caption{Prompts used in the initialization, crossover, and mutation of AEL for TSP: \textcolor{brown}{A description of task}, \textcolor{blue}{Parent algorithm(s)}, \textcolor{darkgreen}{Prompt-specific hints}, \textcolor{purple}{An expected output}, and \textcolor{red}{Other hints}.}
    \label{fig:prompt}
\end{figure}

The details are illustrated in Fig.~\ref{fig:prompt}. The five different colors represent five different components including \textcolor{brown}{A description of task}, \textcolor{blue}{Parent algorithm(s)}, \textcolor{darkgreen}{Prompt-specific hints}, \textcolor{purple}{An expected output}, and \textcolor{red}{Other hints}.

The task involves developing a new strategy for selecting the next node at each step, which remains consistent across all three prompts. Only crossover and mutation include parent algorithms in the prompts. Regarding prompt-specific hints, we instruct the LLM to create a completely new algorithm during initialization, while in crossover and mutation, the algorithm should be inspired by and modified from the parent algorithm(s). This description aligns with the respective functions of each component in the evolutionary framework: the desire for diverse algorithms in the initial population and the expectation that newly created algorithms during evolution will inherit certain aspects from their parents. The format of the expected output remains almost identical for all three prompts. We explicitly define the name, input, and output of the code block for easy identification by the AEL framework. Additionally, we include other hints to emphasize innovation and discourage the need for extra explanations for efficiency and robustness.

\subsection{Experiments}

\subsubsection{Experimental Settings}
The experiments are carried out on 64 randomly generated TSP50 instances, i.e., each algorithm is evaluated on 64 TSP50 instances and the fitness value is the average gap to the optimal solution generated by Gurobi.
The experimental settings for AEL are as follows:
\begin{itemize}
    \item Population size $N$: 10
    \item Number of population $N_g$: 10
    \item Probability for crossover $\sigma_1$: 1.0
    \item Probability for mutation $\sigma_2$: 0.2
    \item Number of parent individuals $l$: 2
    \item Number of offspring individuals $s$: 1
    \item LLM: GPT-3.5-turbo and GPT-4
\end{itemize}

We compared our AEL method to three kinds of existing algorithm design approaches. Note that there are countless works using other frameworks. We compare the methods that follow the same step-by-step constructive framework as that used for AEL. We are able to further promote the performance by applying AEL in other advanced frameworks. 

The three compared approaches as well as the methods are:

\begin{itemize}
    \item \textbf{Algorithm design by humans:} Greedy search (Greedy), which selects the nearest node as the next node. 
    \item \textbf{Algorithm design using domain model:} Neural combinatorial optimization (Domain model), which trains a neural network to learn the heuristic for selecting the next node. In this study, we adopt POMO~\cite{kwon2020pomo}, a widely employed neural solver baseline. We train it on TSP50 using exactly the same settings as in the original paper ~\cite{kwon2020pomo}. The training costs about four days.
    \item \textbf{Algorithm design using LLM:} LLM with prompt engineering (LLM). We directly generate a novel algorithm by instructing LLM. The prompt is the same as that used in the initialization stage of AEL.
    \item \textbf{AEL:} Our proposed AEL.
\end{itemize}

\subsubsection{Experimental Results}

\begin{figure}[t]
    \centering
    \includegraphics[width=\linewidth]{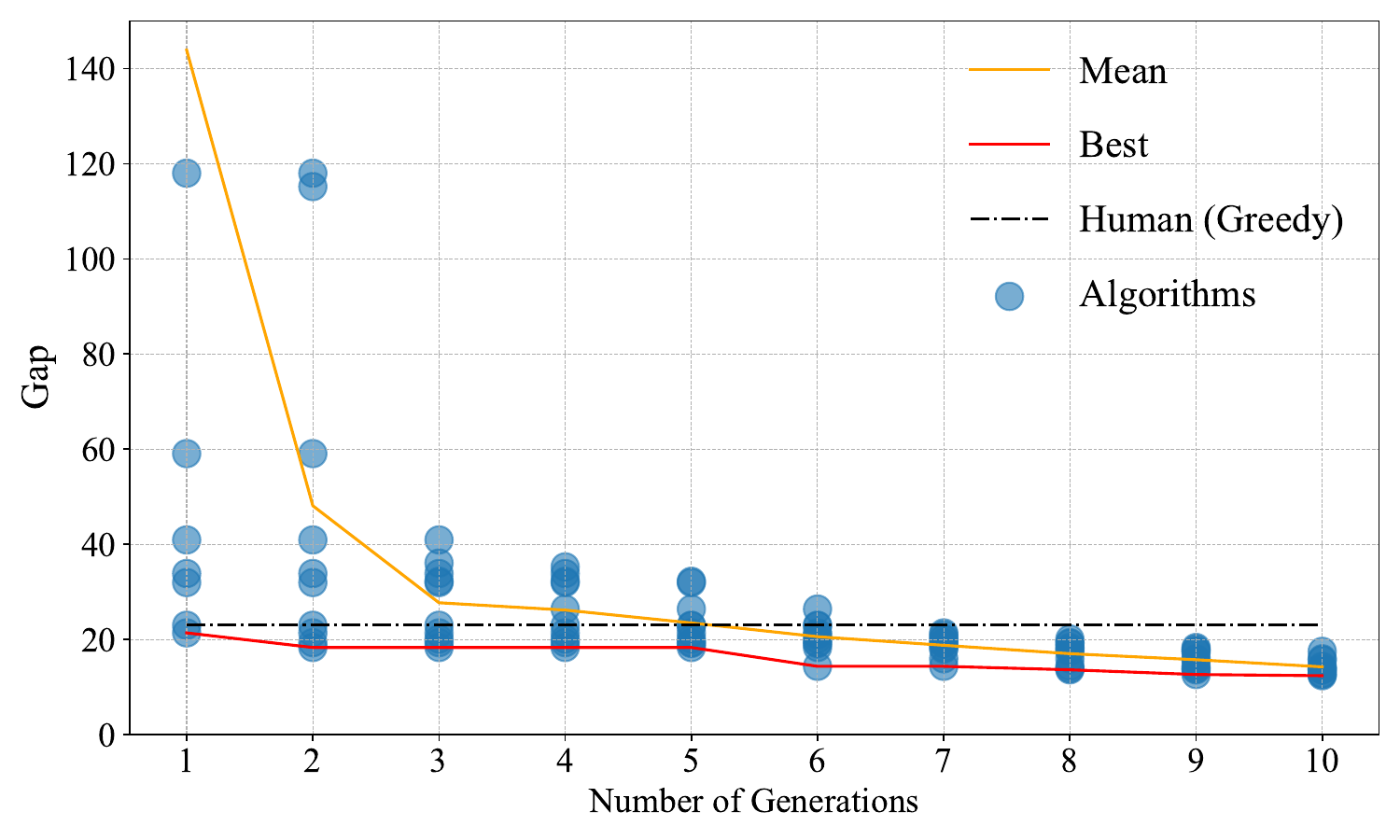}
    \caption{The convergence curve of AEL using GPT-4 on TSP50, where each sample represents an algorithm created in the evolution. The orange and red lines represent the mean and best objective values and the dotted black line represents the greedy algorithm.}
    \label{fig:convergence}
\end{figure}

Fig~\ref{fig:convergence} illustrates the convergence process of the proposed AEL using GPT-4 on the TSP50 dataset. The y-axis represents the gap (\%) to the optimal solution and the x-axis represents the number of generations. Each blue data point represents an algorithm created by AEL during the evolution. The orange and red lines depict the convergence curves of the mean and best objective values, respectively, in each population. The black line represents the greedy algorithm designed by humans. 

It can be observed that there is a clear convergence in terms of both the mean and best objective values as the evolution progresses. AEL nearly converges in 10 generations, and the optimal gap is reduced from 20\% to around 12\%. The best algorithm generated by GPT-4 in the first population is close to the greedy algorithm, while AEL clearly beats the greedy algorithm towards the end of evolution.

\begin{figure}[htbp]
    \centering
    \includegraphics[width=\linewidth]{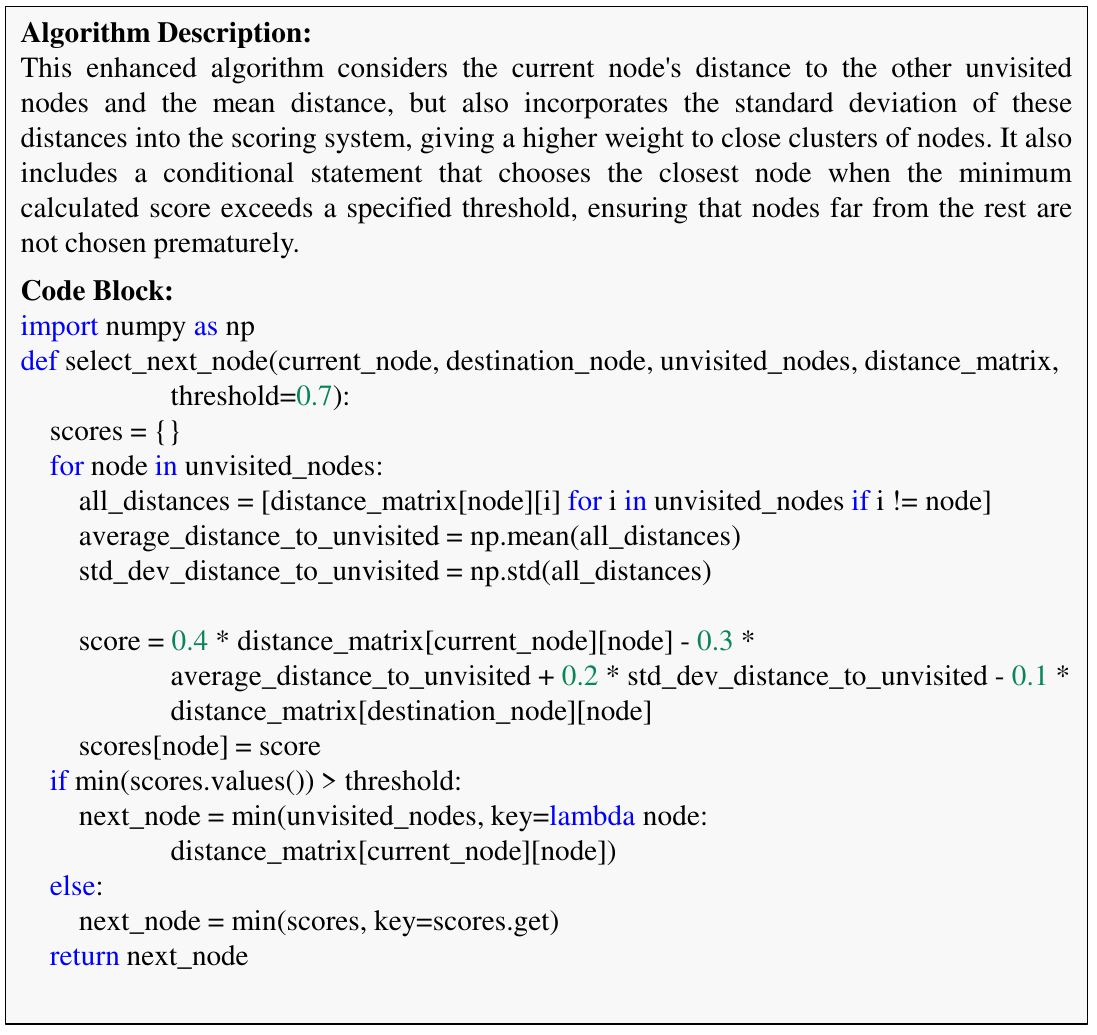}
    \caption{The best algorithm created by AEL using GPT-4. \textbf{Algorithm Description} is a brief algorithm description in two sentences. \textbf{Code Block} includes a Python function named "select\_next\_node" with a pre-defined input and output.}
    \label{fig:bestalgorithm}
\end{figure}

As illustrated in Fig.~\ref{fig:bestalgorithm}, the best algorithm created by AEL is significantly more complicated than the greedy algorithm. It selects the next node considering various factors such as its distance to other unvisited nodes, mean distance, and standard deviation of these distances. The algorithm assigns a higher weight to close clusters of nodes by incorporating the standard deviation into the scoring system. Additionally, the algorithm includes a conditional statement that ensures nodes far from the rest are not chosen prematurely by selecting the closest node when the minimum calculated score exceeds a specified threshold. The created Python code block employs the NumPy library for calculations. The code starts with defining the select\_next\_node function, which takes the current\_node, destination\_node, unvisited\_nodes, distance\_matrix, and an optional threshold parameter. Inside the function, a dictionary called scores is created to store the scores for each unvisited node. The algorithm iterates through each node in the unvisited\_nodes list and calculates the score based on the provided formula. The minimum score is then compared to the specified threshold, and based on the result, the next\_node is determined. Finally, the function returns the selected next\_node.

The automatically designed best algorithm by AEL presents a sophisticated strategy incorporating an additional threshold parameter (not explicitly given in the prompt) and a complex scoring function. The scoring function integrates multiple factors along with four hyperparameters, which presents a challenge, even for an expert, without substantial trial-and-error testing. In contrast, AEL automatically develops the algorithm through an iterative process involving 100 interactions with LLM.

\subsubsection{Evaluation of Optimized Algorithm}
We evaluate the best algorithm designed by AEL  on various TSP instances with multiple problem sizes and compared the results to other algorithm design approaches. Table~\ref{table:results} shows the performance of algorithms designed using different approaches on TSP20 to TSP1000. The results are averaged on 64 randomly generated instances. Apart from the total distance, we also measured the average gap in relation to the baseline solver LKH3. Fig.~\ref{fig:comparison} provides a more intuitive comparison of the average gap vs. the problem size of different algorithm design approaches.

\begin{table*}[htbp]
\centering
\caption{Evaluation of algorithms designed by different approaches on TSP20-TSP1000.}~\label{table:results}
\resizebox{\textwidth}{!}{%
\renewcommand{\arraystretch}{1.2}
\large
\begin{tabular}{llcccccccccccc}
\hline
\hline
\multicolumn{2}{c}{\multirow{2}{*}{Algorithms}}      & \multicolumn{2}{c}{20}           & \multicolumn{2}{c}{50}           & \multicolumn{2}{c}{100}          & \multicolumn{2}{c}{200}          & \multicolumn{2}{c}{500}          & \multicolumn{2}{c}{1000}         \\
\multicolumn{2}{c}{}           & \multicolumn{1}{l}{Dis.} & \multicolumn{1}{l}{Gap.} & \multicolumn{1}{l}{Dis.} & \multicolumn{1}{l}{Gap.} & \multicolumn{1}{l}{Dis.} & \multicolumn{1}{l}{Gap.} & \multicolumn{1}{l}{Dis.} & \multicolumn{1}{l}{Gap.} & \multicolumn{1}{l}{Dis.} & \multicolumn{1}{l}{Gap.} & \multicolumn{1}{l}{Dis.} & \multicolumn{1}{l}{Gap.} \\
\hline
\multicolumn{2}{c}{Baseline (SOTA   Solver LKH3)} & 3.84  & /     & 5.69  & /     & 7.77  & /     & 10.73 & /     & 16.56 & /     & 23.08 & /     \\
\multicolumn{2}{c}{Human (Greedy)}     & 4.49  & 17.0\%& 7.01  & 23.1\%& 9.84  & 26.6\%& 13.50 & 25.8\%& 20.87 & 26.0\%& 28.90 & 25.2\%\\
\multicolumn{2}{c}{Domain Model}                  & \textbf{3.86}            & \textbf{0.6\%}           & \textbf{5.71}            & \textbf{0.4\%}           & \textbf{8.01}            & \textbf{3.0\%}           & 13.02 & 21.3\%& 24.34 & 47.0\%& 36.53 & 58.3\%\\
\hline
LLM (Average)            & GPT-3.5-turbo          & 5.04  & 31.3\%& 7.56  & 32.8\%& 10.62 & 36.7\%& 14.35 & 33.7\%& 22.04 & 33.1\%& 30.05 & 30.2\%\\
      & GPT-4                & 7.45  & 94.2\%& 14.97 & 162.9\%                  & 24.87 & 220.1\%                  & 37.36 & 248.0\%                  & 61.73 & 272.8\%                  & 86.26 & 273.8\%                  \\
LLM (Best)               & GPT-3.5-turbo          & 4.61  & 20.3\%& 6.71  & 17.9\%& 10.01 & 28.9\%& 13.31 & 24.0\%& 20.83 & 25.8\%& 28.98 & 25.6\%\\
      & GPT-4                & 4.36  & 13.6\%& 6.82  & 19.7\%& 9.95  & 28.1\%& 13.94 & 29.9\%& 22.07 & 33.3\%& 30.36 & 31.6\%\\
AEL (Ours)               & GPT-3.5-turbo          & 4.26  & 11.2\%& 6.65  & 16.8\%& 9.32  & 20.0\%& 13.07 & 21.8\%& 20.38 & 23.1\%& 28.34 & 22.8\%\\
AEL (Ours)               & GPT-4                & 4.07  & 6.2\% & 6.33  & 11.1\%& 8.58  & 10.5\%& \textbf{11.94}           & \textbf{11.2\%}          & \textbf{18.67}           & \textbf{12.8\%}          & \textbf{26.03}           & \textbf{12.8\%}     \\
\hline
\hline
\end{tabular}%
}

\end{table*}

\begin{figure*}[htbp]
    \centering
    \subfloat[TSP100, Greedy]{\includegraphics[width=0.33\linewidth]{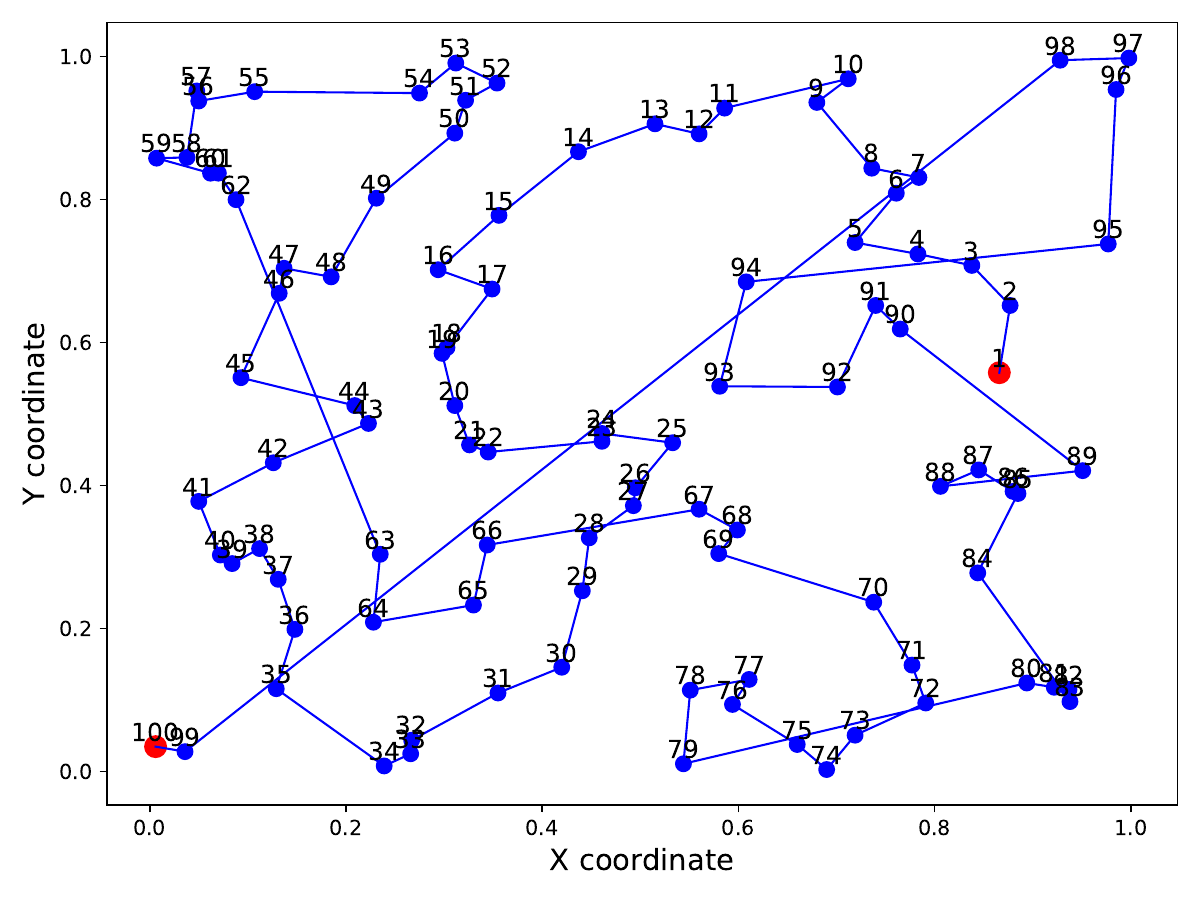}}
    \subfloat[TSP100, AEL (GTP-3.5-turbo)]{\includegraphics[width=0.33\linewidth]{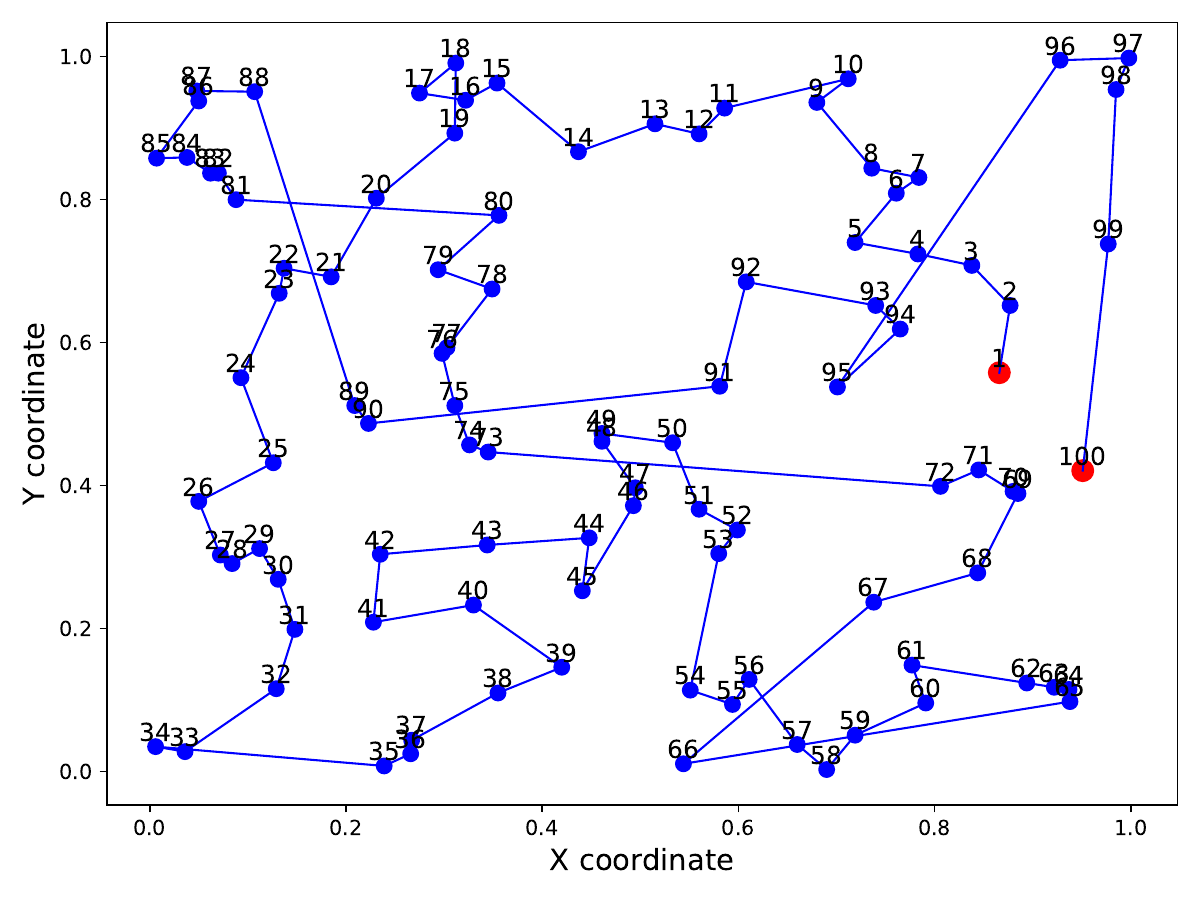}}
    \subfloat[TSP100, AEL (GTP-4)]{\includegraphics[width=0.33\linewidth]{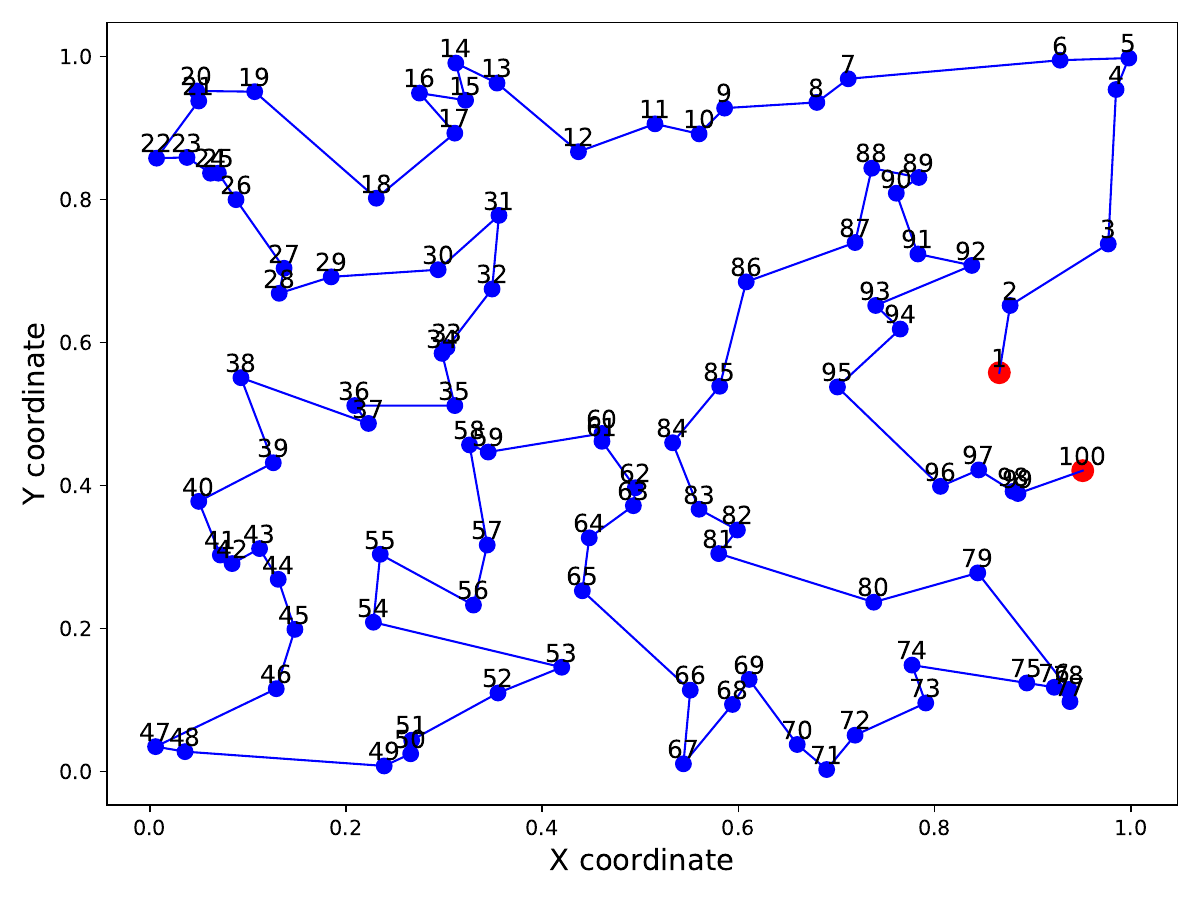}}\\
        \vspace{-5pt} 
    \subfloat[TSP500, Greedy]{\includegraphics[width=0.33\linewidth]{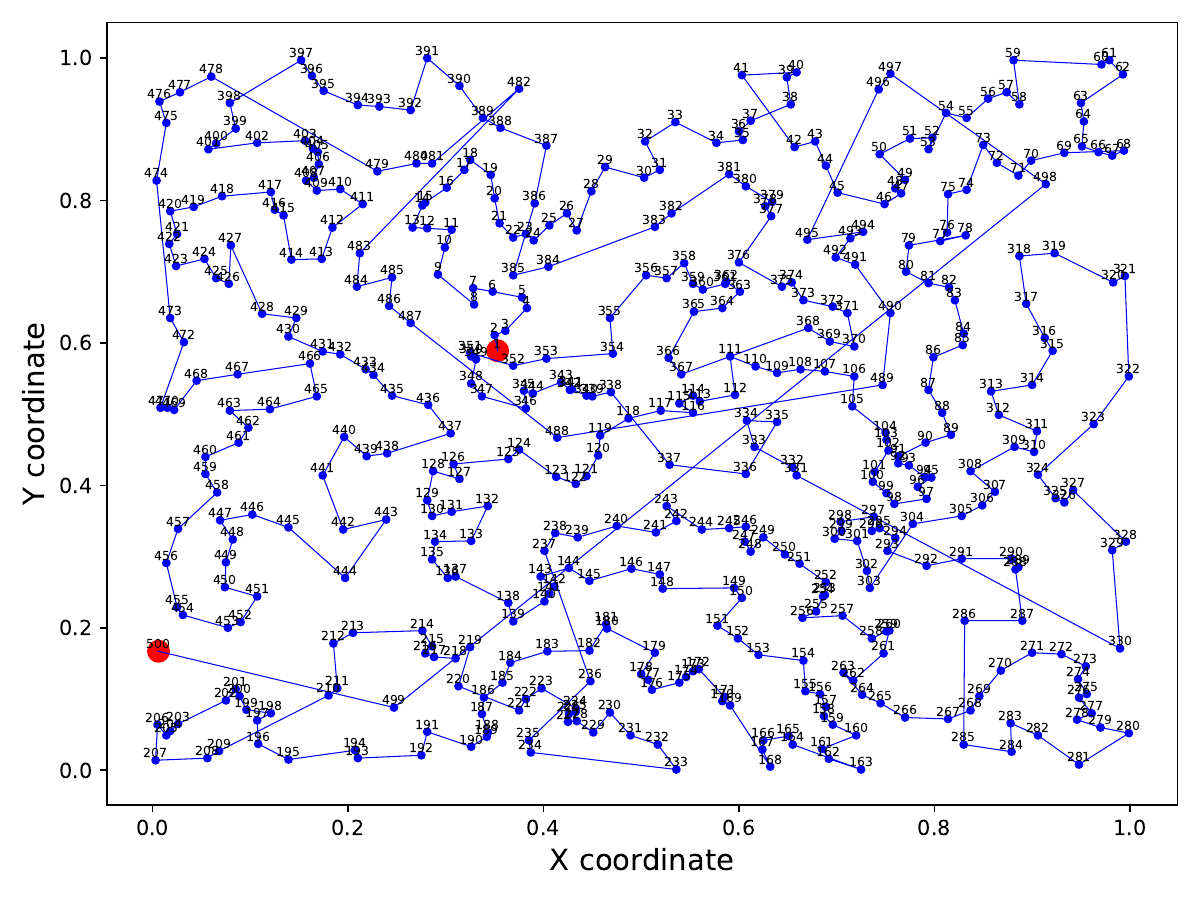}}
    \subfloat[TSP500, AEL (GPT-3.5-turbo)]{\includegraphics[width=0.33\linewidth]{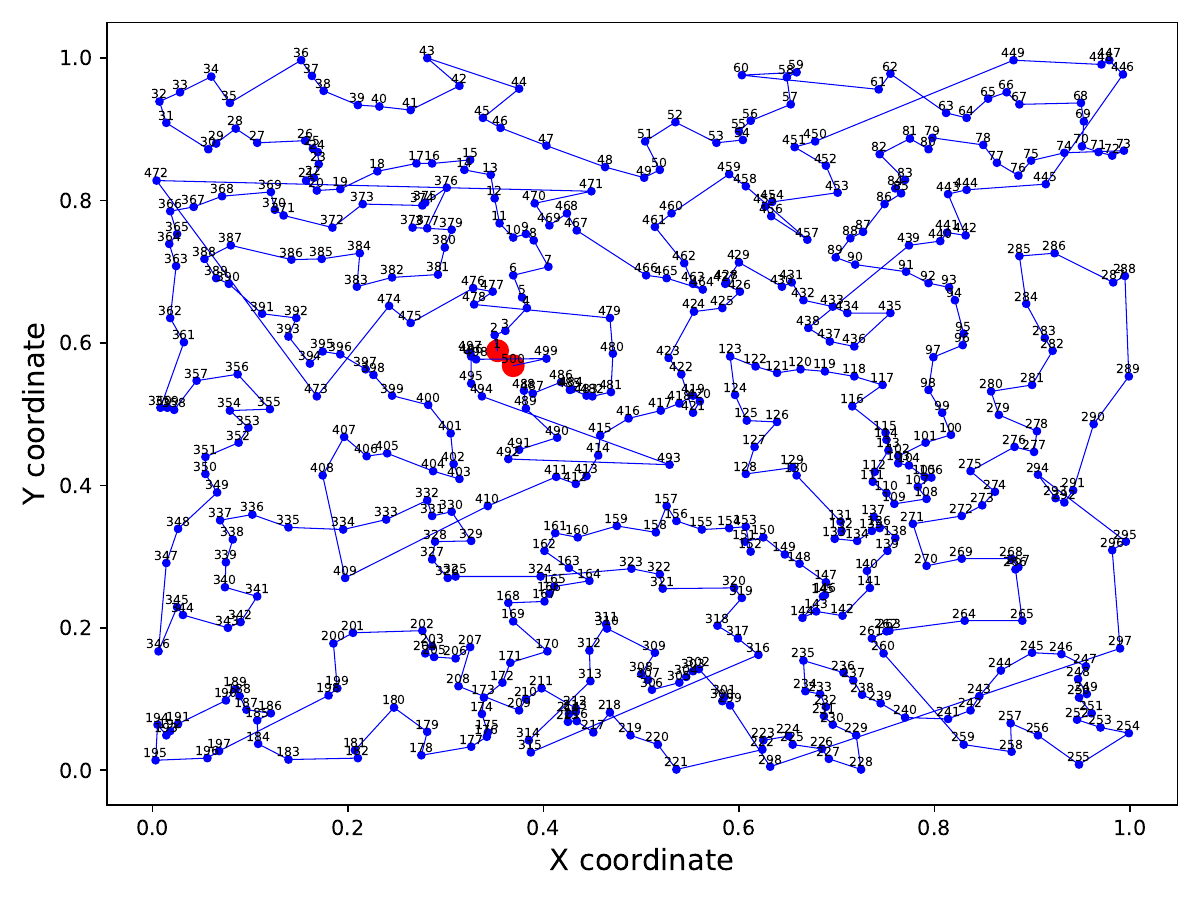}}
    \subfloat[TSP500, AEL (GPT-4)]{\includegraphics[width=0.33\linewidth]{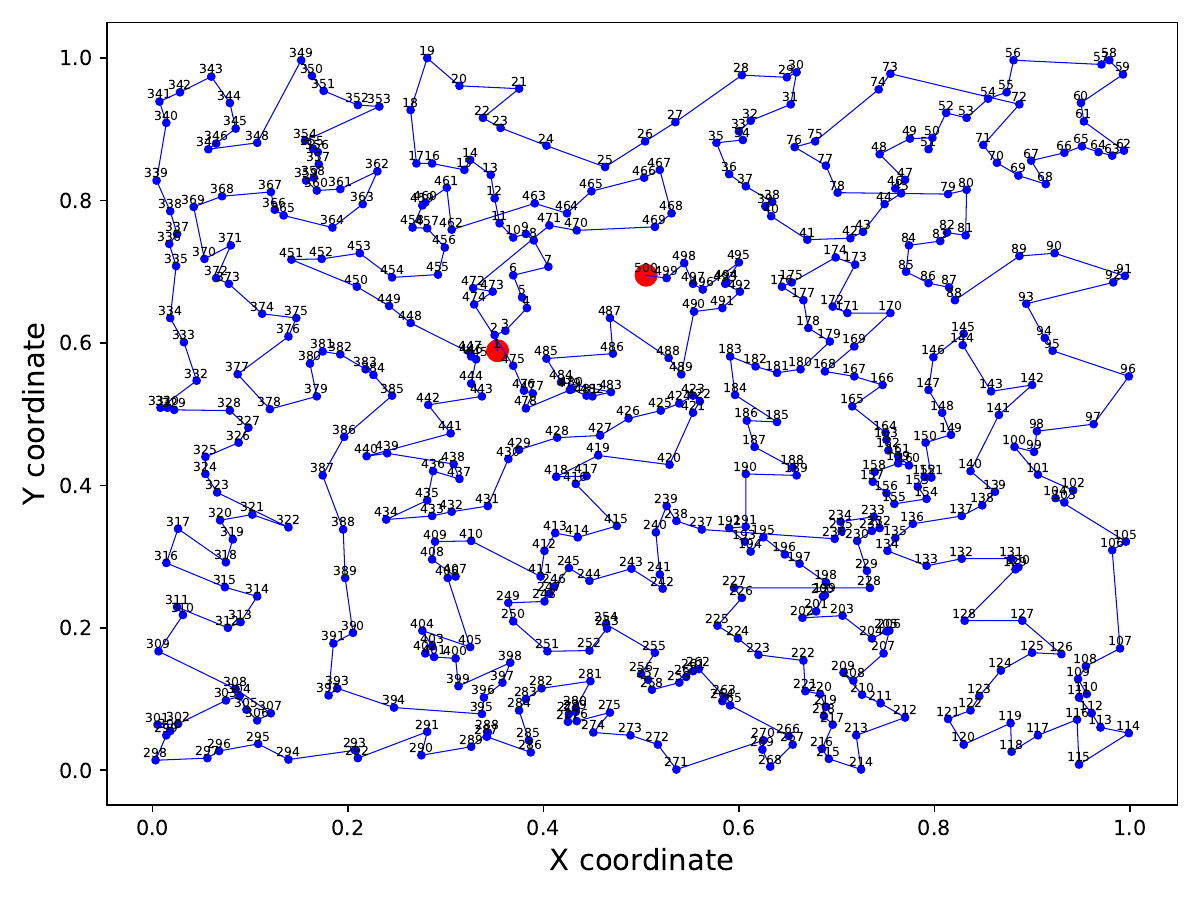}} \\
        \vspace{-5pt} 
    \subfloat[TSP1000, Greedy]{\includegraphics[width=0.33\linewidth]{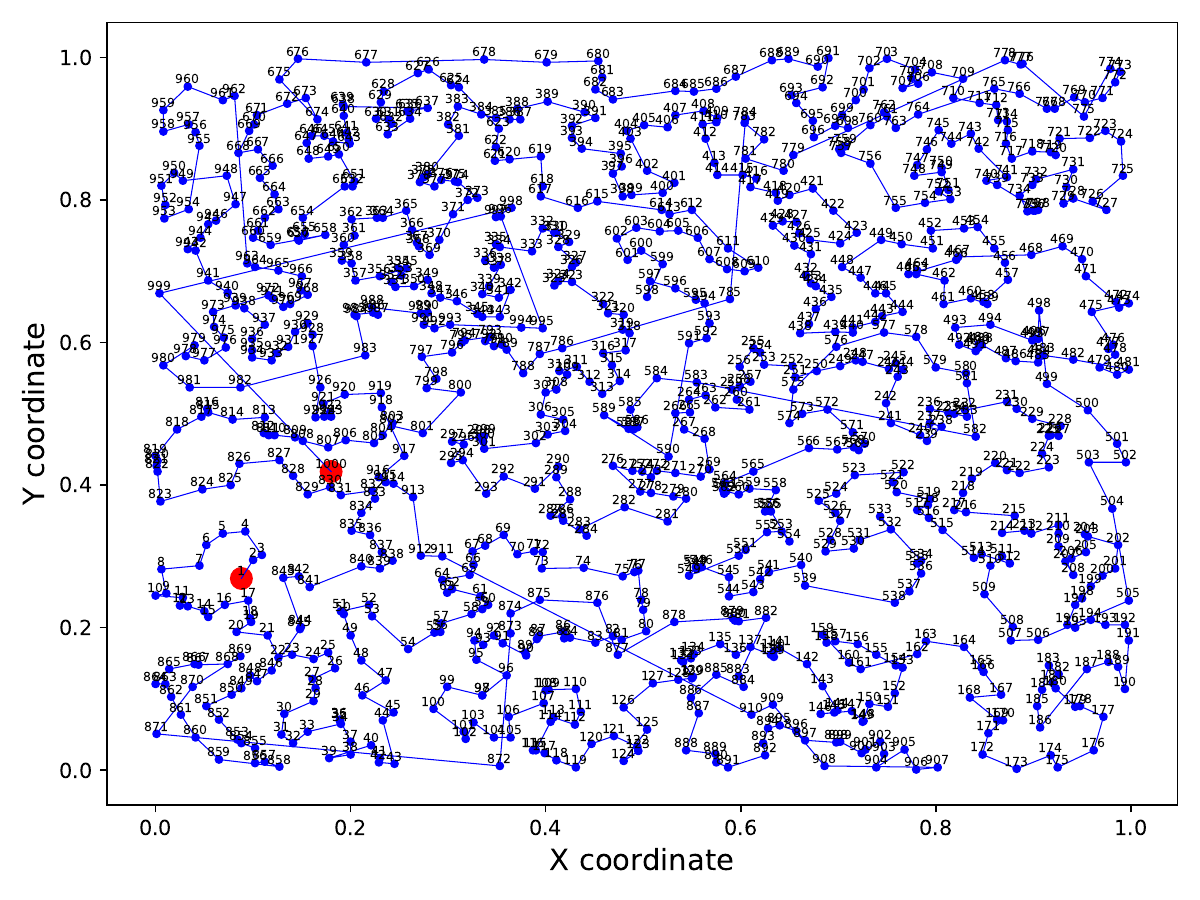}}
    \subfloat[TSP1000, AEL (GPT-3.5-turbo)]{\includegraphics[width=0.33\linewidth]{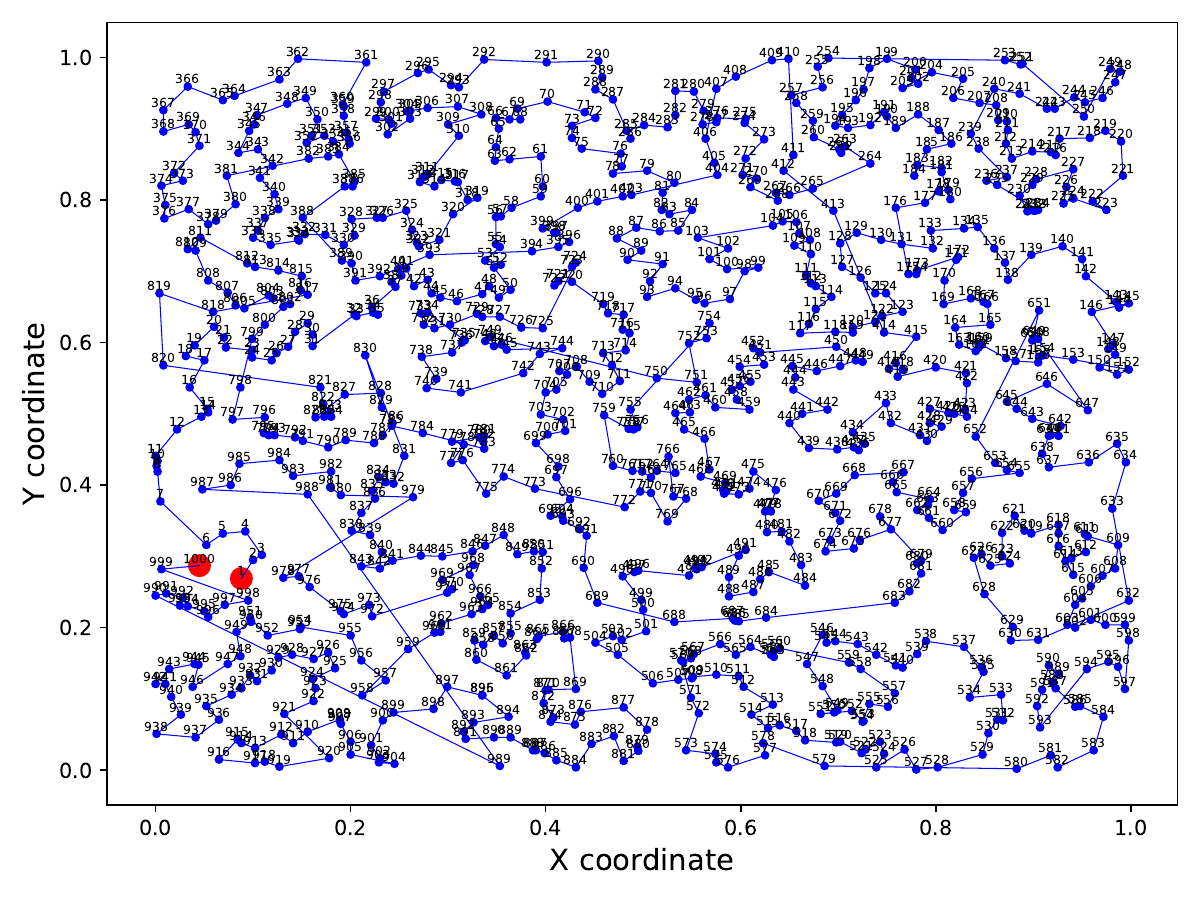}}
    \subfloat[TSP1000, AEL (GPT-4)]{\includegraphics[width=0.33\linewidth]{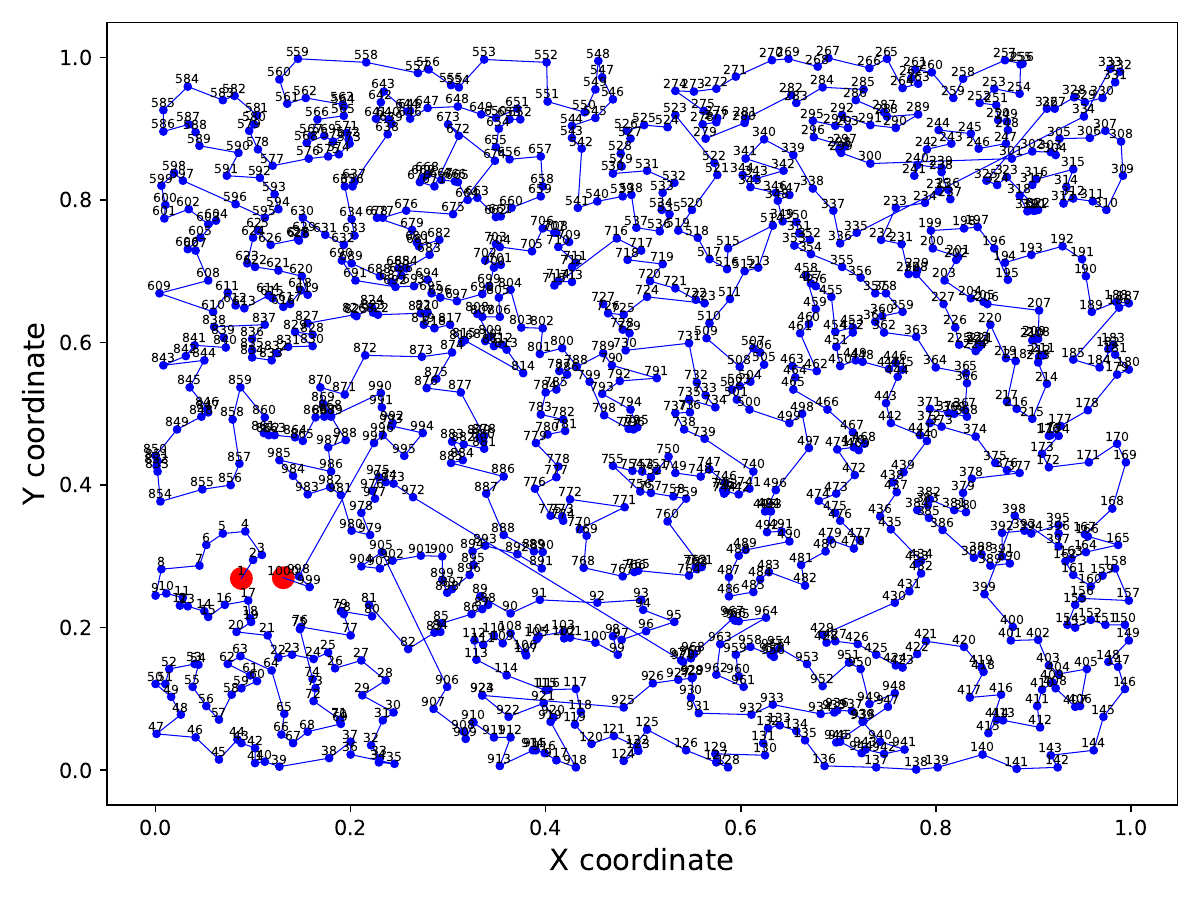}}
    \caption{A comparison of the greedy algorithm and algorithms developed by AEL, utilizing GPT-3.5-turbo and GPT-4, for solving TSP100, TSP500, and TSP1000 instances. The nodes in the scatter plot represent locations, while the blue lines represent routes generated by the various algorithms. The numbers displayed above the scatter plot indicate the sequence in which the locations appear in the route. The starting and ending locations are denoted by red nodes.}
    \label{fig:route}
\end{figure*}

\begin{itemize}
    \item AEL outperforms the simple Greedy algorithm designed by humans in all problem sizes. The average gap is reduced by half from 17.0\% to  6.2\% and from 25.2\% to 12.8\% for TSP20 and TSP1000, respectively.
    \item AEL demonstrates significantly better generalization performance across various problem sizes when compared to the domain model. The domain model is trained and overfitted on TSP50, whereas AEL presents a much more robust solution. Although the domain model surpasses AEL in problem sizes close to the training data, its performance rapidly deteriorates on large-scale problems. The average gap increases dramatically from less than 1\% to over 50\%.
    \item AEL also outperforms directly instructing LLM to design algorithms. LLM (Average) and LLM (Best) represent the average and best of ten algorithms directly generated by instructing LLM. The best LLM algorithm is competitive to the greedy algorithm but evidently inferior to AEL.  Interestingly, the best gap achieved by LLM with GPT-4 is inferior to GPT-3.5-turbo. An explanation for this is that GPT-4 is more powerful but can also be overly innovative with excessive randomness. Consequently, the algorithms directly generated by GPT-4 could be considerably worse.
    \item We also want to note that the demonstration is conducted based on a basic constructive heuristic framework. Our comparison involves the algorithm created by AEL, a greedy algorithm, and a domain model, all using the same step-by-step constructive framework. However, there are numerous other complex algorithms capable of generating near-optimal solutions for TSP~\cite{pan2023h}. Additionally, recent neural solvers specifically designed for large-scale TSP have also demonstrated good generalization performance~\cite{zhou2023towards,cheng2023select,luo2023neural}. Advanced frameworks can be integrated to promote performance in the future.
\end{itemize}

Fig~\ref{fig:route} compares the routes generated using the greedy algorithm and algorithms developed by AEL, utilizing GPT-3.5-turbo and GPT-4, on TSP100, TSP500, and TSP1000 instances. The scatter plot nodes represent locations, with the blue lines indicating the routes generated by different algorithms. The numbers displayed above the scatter plot indicate the sequence in which the locations appear in the route. The starting and ending locations are marked in red. The results demonstrate that the algorithms designed by AEL produce superior routes with fewer intersections than the greedy algorithm, resulting in shorter total distances. Additionally, the starting and ending nodes are closer together compared to the greedy algorithm. AEL using GPT-4 outperforms AEL using GPT-3.5-turbo.


\subsection{Created Algorithms or Spliced Algorithms}
It is debatable whether the LLM can genuinely comprehend and generate new knowledge or if it simply searches and combines various existing information~\cite{bubeck2023sparks}. As the TSP is a well-studied combinatorial optimization problem with numerous publicly available resources and research papers, it is possible that the LLM merely selects or splices existing algorithms~\cite{cook2011traveling}. However, the majority of the algorithms created in our AEL framework cannot be found on the web or in publications. The comparison between AEL algorithms and LLM algorithms also demonstrates that AEL stimulates the creation of novel algorithms to a noteworthy extent by combining evolutionary computing and LLM.

\subsection{Less/No Expert Knowledge}
AEL requires minimal expert knowledge about the target problem. For designers using our AEL framework, the primary workload lies in designing the prompt engineering for each component for the target problem. The prompt engineering is presented in a natural language format, making it accessible to algorithm designers from diverse backgrounds.

\section{Future Works}\label{sec:future}

\begin{itemize}
    \item \textbf{AEL with tools:}  Equipping LLMs with external tools significantly enhances the capabilities of the model~\cite{mialon2023augmented,schick2023toolformer,paranjape2023art}. There are numerous well-crafted existing optimization algorithms, which can be integrated as effective tools. This approach provides a more flexible and robust framework as opposed to relying solely on LLM for algorithm evolution. We can also integrate other heuristic frameworks into AEL. For example, landscape updating for guided local search~\cite{hudson2021graph}, improve large neighborhood search~\cite{hottung2019neural}, and Tabu search~\cite{nguyen2020deep}.

    \item \textbf{AEL with additional information:} Another interesting direction is to provide additional information as the input for LLM during the optimization process. The information can be in the form of history search trajectories, external archives, and rewards obtained during optimization~\cite{jiang2022evolutionary,ishibuchi2023new}. With more available information, AEL is able to evolve more powerful algorithms.

    \item \textbf{AEL for complex optimization problems:} Complex problems pose challenges for AEL as they are difficult for LLMs to comprehend, which is even challenging for humans. In such cases, it is possible to decompose the problem into simpler tasks and adopt LLMs for each task~\cite{khot2022decomposed}. A possible solution is to separate the algorithm description and coding and evolve in a hierarchical way, i.e., evolve on the algorithm and then generate code for each algorithm. Refinement techniques~\cite{chen2023teaching,press2022measuring,wang2022self} can be applied to reduce the failure rate.

    \item \textbf{Multi-objective AEL:} 
    The current focus of AEL is mainly on single-objective optimization problems. However, many real-world problems are multi-objective in nature, where multiple conflicting objectives need to be simultaneously considered. As AEL is algorithm-agnostic, we can easily extend AEL to handle multi-objective optimization problems using multi-objective optimization algorithms~\cite{zhou2011multiobjective}. Multi-objective AEL finds trade-off algorithms that satisfy multiple objectives, which can be particularly useful in real-world decision-making scenarios~\cite{bezerra2015automatic}.
    
\end{itemize} 

\section{Conclusion}\label{sec:conclusion}



This paper introduces a novel approach called Algorithm Evolution with Large Language Model (AEL) for automatic algorithm design. By utilizing a large language model (LLM), AEL automates the generation of optimization algorithms using an evolutionary framework. It significantly reduces the need for expertise knowledge and domain model training.

We have demonstrated the effectiveness of AEL on the constructive method for TSP. Results on TSP instances with problem sizes ranging from 20 to 1000 show that the algorithm generated by AEL outperforms the simple hand-crafted greedy algorithm and the algorithms generated by directly instructing LLMs. It also exhibits excellent scalability across different problem sizes compared to training a domain neural model. Future works on the integration of AEL with more advanced algorithm frameworks could lead to even more powerful algorithms.

\bibliography{AEL.bib}
\bibliographystyle{IEEEtran}

\end{document}